# Emotion Detection in Twitter Messages Using Combination of Long Short-Term Memory and Convolutional Deep Neural Networks

B. Golchin, N. Riahi

**Abstract**—One of the most significant issues as attended a lot in recent years is that of recognizing the sentiments and emotions in social media texts. The analysis of sentiments and emotions is intended to recognize the conceptual information such as the opinions, feelings, attitudes and emotions of people towards the products, services, organizations, people, topics, events and features in the written text. These indicate the greatness of the problem space. In the real world, businesses and organizations are always looking for tools to gather ideas, emotions, and directions of people about their products, services, or events related to their own. This article uses the Twitter social network, one of the most popular social networks with about 420 million active users, to extract data. Using this social network, users can share their information and opinions about personal issues, policies, products, events, etc. It can be used with appropriate classification of emotional states due to the availability of its data. In this study, supervised learning and deep neural network algorithms are used to classify the emotional states of Twitter users. The use of deep learning methods to increase the learning capacity of the model is an advantage due to the large amount of available data. Tweets collected on various topics are classified into four classes using a combination of two Bidirectional Long Short Term Memory network and a Convolutional network. The results obtained from this study with an average accuracy of 93%, show good results extracted from the proposed framework and improved accuracy compared to previous work.

## I. INTRODUCTION

NOWADAYS, the users' opinions in any of the social media are very important and they are effective in many cases [1]. On news sites, different opinions are written by users about a news or person. The existence of difference in opinions not only reflects the interests of users, but it can also reflect the opinions of each user. For example, different opinions of users in online stores can indicate the level of customers' satisfaction and product quality, and can also be a good guide for other buyers. Stores can provide a good platform for users to shop online displaying users' feedback. On the other hand, because users' opinions can express their opinions, these are very important for governments and political and social institutions. In essence, social media has emerged as a popular medium for sharing ideas, thoughts, information, and sharing useful experiences. In many cases, users' emotional and spiritual states can be extracted using their opinions on social media.

Classifying these emotional states can provide useful information such as adapting the content to each person's emotions, marketing campaigns, monitoring the responses to local and public events, and discovering the individual state trends [2]-[4].

One of the significant usages of emotion classification is the extent and model of people's reactions during crises. For example, in the event of a social crisis, a large number of users share textual content on social networks such as Twitter. Content produced during crises is divided into two categories: noise or unused content and useful content for recognizing emotional states. The extent of emotional involvement in a crisis can be determined by using useful content. Understanding the scope of the crisis or understanding the details of the scope of the crisis are pieces of data that can be used to raise situational awareness [5], [6].

Reference [6] categorizes the people's emotions on Twitter during the sandstorm crisis. In this research, gathering appropriate data and analyzing them are two most important and main parts of this research. Then, according to the categorization of user messages, crisis managers perform appropriate user-centered activities to understand the needs of the user and provide more information about social media tools applied to manage crises using the initial results. An important issue here during crisis management is the possibility of distinguishing between negative emotions such as fear of anger or distinguishing between different types of positive emotions such as happiness and excitement, so it is necessary to use not only users' emotions as positive and negative but be divided into smaller categories.

Reference [7] determines the human emotion during COVID-19 pandemic by using various machine learning approaches. In this paper, eight emotions such as anger, anxiety, desire, disgust, fear, happiness, relaxation and sadness are determined.

In [8], emotions expressed in a large collection of cancer-related tweets demonstrates that joy was the most commonly shared emotion, followed by sadness and fear, with anger, hope, and bittersweet being less shared. Furthermore, both the gatekeepers and influencers were more likely to post content with positive emotions, while gatekeepers refrained themselves

B. Golchin is with the Alzahra University, North Sheikh Bahaee St., Deh-e Vanak, Tehran, I.R. of Iran (e-mail: bahar.golchin72@gmail.com).

N. Riahi, is with the Department of Computer Science, Alzahra University, North Sheikh Bahaee St., Deh-e Vanak, Tehran, I.R. of Iran (e-mail: nriahi@alzahra.ac.ir).

from posting negative emotions to a greater extent. Last, cancer-related tweets with joy, sadness, and hope received more likes, whereas tweets with joy and anger were more retweeted.

Identifying suicidal content from tweets based on emotional sentiment has been addressed in [9]. According to the findings, tweets about suicide exclusively expressed fear, sadness, and issues of negative sentiments. So, using the emotional awareness classification system, the researchers found emotion plays a key role in facilitating online suicide-related content detection.

As a result, the expression of feelings and emotions in social networks can have a great impact on decisions related to different institutions, and as described in the previous paragraph, we can examine the reactions of people during different events according to their opinions on social networks and we can use these analyzes for the purposes of this process [6], [10].

The transmission of users' opinions on various issues in social networks such as Facebook and Twitter, has increased greatly with the development of social networks and the existence of minimum restrictions on these networks to share text and speech messages. Analyzing emotions from the text documents seems to be challenging since textual expressions are not always directly use the emotion related words, but often outcome from the understanding of the meaning of concepts and interaction of concepts mentioned in the text document [11]. Detecting users' emotions and categorizing their emotions can have profound effects on management, political science, economics, health, and social sciences [12], [13].

## II. PROPOSED APPROACH

The proposed data set model, as data extracted from the Twitter social network, is labeled based on the hashtag and defined as Emotex in [14], then the preprocessing is done proportional to the data form and the data is tokenized, and the numerical vector is extracted by different methods, such as the Glove embedding vector, and it is used as input to the deep neural network desired in the article [15], [16]. The desired emotions are classified after training the deep neural network to obtain the best weights and finally to use them in the test data set. In this work, as mentioned, the social network "Twitter" and English-language tweets have been used with various issues.

The most important part of emotion classification is getting the right accuracy when using a large data set. Deep learning methods [17] are an advantage when the data set is large, because the learning is done on more parameters, and therefore deep learning methods have more learning capacity on the model. And when a function has many parameters, it can certainly fit more complexities. Thus, innovation in this work includes the use of a combination of two deep convolution neural networks [18] and long short-term memory [19] (LSTM-CNN) to increase the accuracy of previous work. The advantage of combining these two deep neural networks (LSTM-CNN) is that, the long short-term memory network has the ability to receive it in the form of a whole sentence-based tokens instead of receiving the input, and since this network has memory, when each token received as input, this memory is updated and a better understanding of that sentence is received over time, which determines the limits of emotional states. This output is then received as the input of the convolution network, and the convolution network uses its own important ability i.e., local feature recognition to identify local and valuable input features, thus the accuracy will be increased.

## III. MODEL OF EMOTIONS

There are two main models for expressing emotions: A. basic model and B. dimensional model. Each of these models helps to cover different aspects of human emotions, and both can provide understanding and insight on how emotions are expressed and interpreted in the human mind.

### A. Basic Emotional Models

The easiest way to identify emotions is to use emoticon words or classification tags. This model assumes that the emotional categories are the same as the Paul Ekman's Six Basic Emotions [20] or the Specific Domain categories. Each emotion is characterized by a set of specific characteristics and conditions of its occurrence. Most work focuses on six basic emotion categories; however, many researchers have used different emotion sets for different areas. For example, the research [21] used five categories of fatigue, confusion, pleasure, freedom, and frustration to describe effective states in a student system. Students seldom felt fear or hatred, and usually they felt tired or happy, which is evidence of the need for domain-based categories.

However, emotion classification models have weaknesses due to the limited number of tags. For example, emotion categories contain discrete elements, and many types of emotions can be observed in each individual category. Categories do not cover all emotions because a large number of emotions can be seen in separate categories. These findings suggest that the emotional categories may not display individual emotional states well, and that we may have unrelated and undesirable diagnoses. This can lead to the problem of mandatory selection of an irrelevant category due to the lack of a relevant category. Thus, the classification model may have limitations in identifying emotional states perceived by individuals [14], [22].

However, this type of classification model is dominant and definitive and due to its simplicity and reputation, many changes have been made in it and it is used optimally. The only way that classification models may be different is the difference in the number of categories in the model [23].

### B. Dimensional Emotional Models

The second approach to identifying emotions is to use a dimensional model in which the effects are shown in a dimensional form. Emotional states in this model are related together by a set of common dimensions and they are generally defined in a two- or three-dimensional space. Each of the emotions occupies a place in this space. Through the dimensional model, the degree of similarity between the categories of emotions can be well recognized, but it is still less used due to its special complexity [21]. In this section, some dimensional models are briefly reviewed.

According to Fig. 1, the circumplex model is represented by a set of points representing the emotions. Emotion-related words are organized in a circumplex form in such a way that they can choose each subject in a place between two words related to discrete emotions. The theory of circumplex effectiveness identifies two main dimensions of positive and negative effects and it considers a high to low scope. Numerical data are obtained from the relative position of points in a bipolar, two-dimensional (Valence-Arousal) space. The valence dimension shows the positive and negative emotions on both sides. Arousal dimension distinguishes calm states versus emotional states [24].

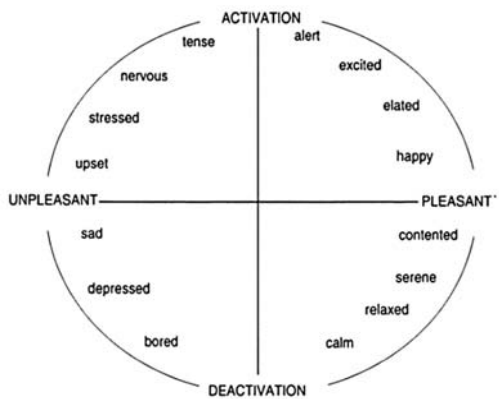

Fig. 1 A graphical representation of the circumplex model

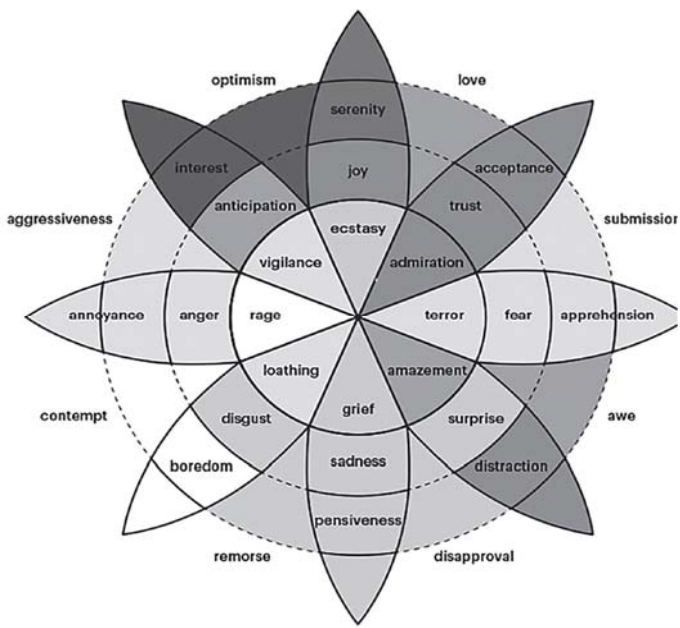

Fig. 2 Plutchik Emotions Cycle

Research [25] has used other angles of the emotion circle to indicate emotion categories. Hence the model is called the emotions circle as shown in Fig. 2. In this research, in addition to six categories of research [20], two categories of trust and expected emotions were added to the set of emotions. These eight emotions are organized into four sets of dipoles. Happiness versus sadness, anger versus fear, trust versus hatred and surprised versus expected. Fig. 2 shows the relationships between emotions [26]. As shown in this figure, emotions like colors can vary in intensity so that proximity to the center indicates the intensity of emotions and they can be combined with each other to create additional emotions.

Research [27] has also proposed a model related to the learning process for emotions in a Valence-Arousal map. Fig. 3 is used in a fully automated computer program to detect learner emotions.

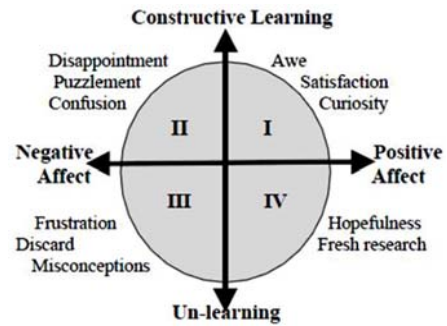

Fig. 3 Cohort Emotions Model

IV. METHODOLOGY

A. Preprocessing

The preprocessing stage is defined and usable in each network according to the type of network and the ultimate goal of the network. In the following, the seven used methods are explained in this research.

1) Tweets often have a username. This username starts with @ symbol before the username (for example @Marilyn). In this stage, we will replace all words that start with @ with the keyword USERID.
2) Since many tweets contain URLs, all of these URLs will be replaced with the keyword URL.
3) Words that contain repetitive letters like happyyyyyyyy are seen a lot in Twitter messages; each letter in the word that has been combined more than twice is replaced by the same letter. For example, the word "happyyyyyyyy" changes to happy.
4) Many tweets have multiple hashtags that may be for different classes, for example the tweet "Got a job interview today with At & t… # nervous #excited" has #nervous hashtags from the class "Unhappy-Active and #excited" from the class "Happy-Active". Tweets containing hashtags from different classes are removed from the training data because these types of tweets cause ambiguity in the training data and mislead the classification algorithm for classifying the classes.
5) Some tweets contain different emotions from different classes. For example, the tweets "Tomorrow, first volleyball match :) and final exam :(" contain both the emotions of happiness and sadness that these tweets are deleted in the pre-processing stage.

6) Some tweets contain incompatibility between hashtags and emotions. For example, the tweet "Yup, I am totally considering leaving this planet now :) #disappointed #nohope", contains the hashtag #disappointed from the Unhappy-Inactive class and feeling :) which indicates happiness. These types of tweets are removed from tagged data.
7) Hashtags will be deleted at the end of the tweet and hashtags in the middle or at the beginning of the tweet will be retained if they are part of a sentence, otherwise they will be deleted. This is because if these tweets are not deleted, a large amount of weights will be adjusted based on hashtags, which will damage accuracy.

One of the most common methods in processing the natural languages is tokening and it can be considered as a kind of preprocessing method. Using the tokenization operation, the text is first received as input and it is divided into separate words. The result from tokenization is used as input to other stages of network analysis. In English, words are separated by a space and are considered a token [28].

*B. Method*

Fig. 4 shows the stages of implementing the proposed framework for classifying Twitter users' emotions using the Deep Learning Neural Network algorithm.

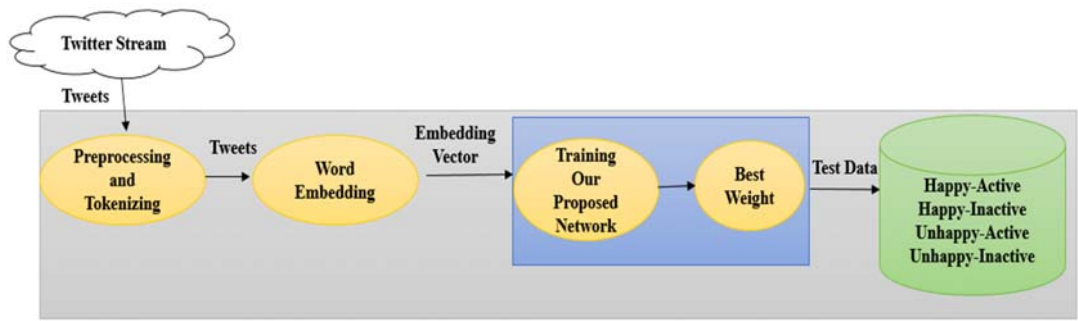

Fig. 4 Stages of implementing the proposed framework

This section presents a proposed method, which is a model based on a deep neural network for detecting and classifying users' emotional states on the Twitter social network. As mentioned before, the proposed model is a combination of two deep neural networks, long short-term memory and convolution (LSTM-CNN).

As shown in Fig. 5, this model includes a long short-term memory input layer that receives word embedding for each token in the tweet as input. Output tokens store not only the information of the input tokens, but also the information of each token in the previous stages. In fact, the long short-term memory layer generates new encryption for the main input. The output of the long short-term memory layer, as the input of the convolution layer, is expected to extract local features, and eventually the output of the convolution layer is reduced to smaller dimensions; and the emotion classification classes of the tweets are specified [29]. The advantage of combining these two deep neural networks is that the long short term memory network has the ability to receive input in the form of tokens instead of receiving a complete sentence, and it acts as an encoder; since this network has a memory and each token is received as input, this memory is updated and a better understanding of that sentence is received over time, which determines the limits of emotional states. Then this output is received as input to the convolution network and the convolutional network uses its important feature, i.e., recognition of local feature detection, to detect local and valuable input features, thus increasing accuracy.

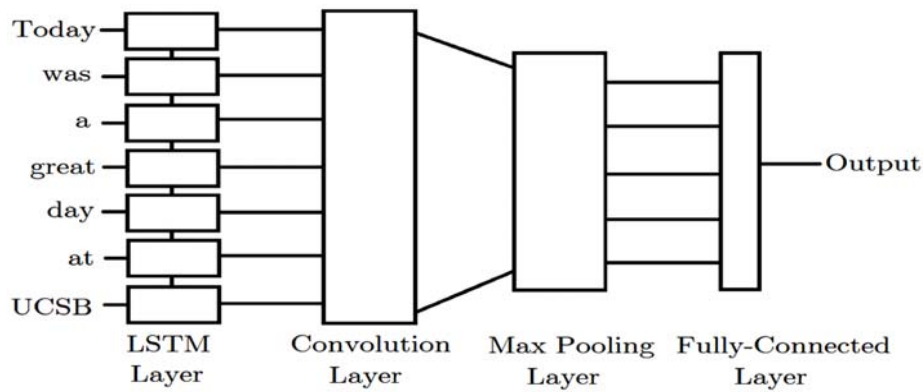

Fig.5 Long short-term memory and convolution (LSTM-CNN)

*C. Evaluation*

The dataset used in this work includes 100,000 tweets as training data on a variety of issues. A total of 33,000 tweets have been used as test data [14], and 20% of the training data was used for validation. In this research, 200 tests have been performed to obtain each of the network parameters. The accuracy, recall, and F-score criteria will be calculated based on the Confusion matrix, and these criteria will be used for comparison with other articles.

*D. Baseline Methods*

In this section, in order to evaluate and measure the results more accurately, we implemented the data set in the reference [14] provided by the author of the article with the proposed method and the obtained results were compared with that available in the mentioned article. Features used in [14] include Unigram, emoticon, Punctuation, and negation attribute.

Also, machine learning methods include three simple Bayesian algorithms and decision tree and backup vector machine are used on the data set of this article to classify the emotional states of Twitter users into four classes Happy-Active, Unhappy-Active, Happy-Inactive, and Unhappy-Inactive. For comparison, the criteria obtained in three experiments were used for each class.

TABLE I
COMPARING THE CLASSIFICATION RESULTS OF OUR PROPOSED METHOD WITH EMOTEX [14] BASED ON PRECISION, RECALL, AND F-MEASURE

| Emotion Classes | Our Proposed Method | | | Emotex | | |
|---|---|---|---|---|---|---|
| | Prec. | Rec. | FM | Prec. | Rec. | FM |
| Happy-Active | 92 | 94 | 93 | 84.2 | 95.4 | 89.5 |
| Happy-Inactive | 93 | 91 | 92 | 94.3 | 84.4 | 89.1 |
| Unhappy-Active | 92 | 92 | 92 | 91.4 | 90.5 | 91 |
| Unhappy-Inactive | 90 | 91 | 90 | 91.2 | 88.4 | 89.8 |

V. EXPERIMENTAL RESULTS

*A. Dataset*

In order to collect data, the social network "Twitter and English tweets" have been used related to this social network. Twitter data is a rich source for receiving information on any conceivable issue. This data can be used for a variety of purposes, including finding trends related to a particular keyword, gathering the feedback on new services and products, and measuring the feelings and emotions. Tweets can be extracted randomly, on different issues and in different formats.

The hashtags on the Twitter social network have been used to collect tweets. A hashtag is a tag used to categorize and share posts and comments on an issue globally and beyond the circle and friends list. A hashtag is exactly a kind of tag that is pasted on various issues on social networks like Twitter so that users can have a collection of information about a particular issue whenever needed. This trend is done using the # sign before the intended word [14].

A large set of tagged messages has been created using hashtags to automatically interpret text messages containing emotions. These messages are then ready to be used as tagged data to teach classifiers. Fig. 6 shows the data collection process. First, a list of emotion hashtags is needed to be defined to collect tagged emotion messages [14].

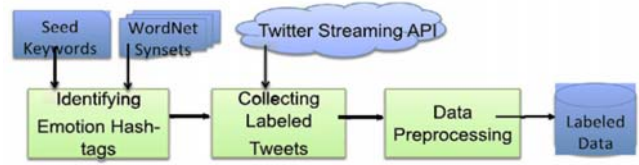

Fig. 6 Data collection process

For this purpose, the circumplex model is used to extract 28 important and key words shown in Fig. 7. It is used as input to a set of keywords and their expansion using WordNet's synsets and similar meanings.

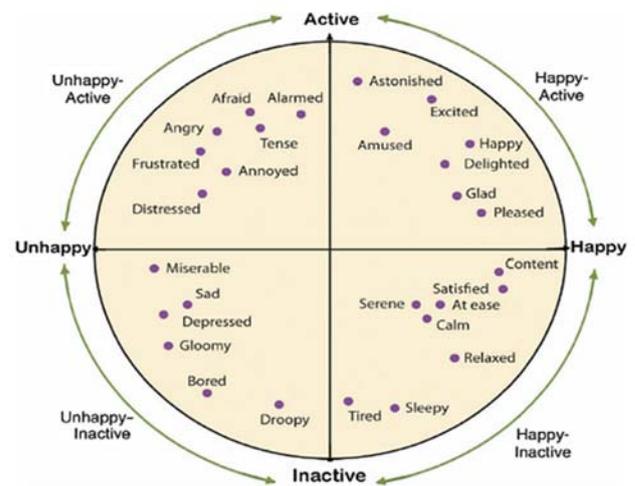

Fig. 7 Circumplex model

An extended set of keywords is used to identify emotion hashtags. Tweets containing one or more hashtags are then gathered from the hashtag collection, thus ensuring that there are tagged tweets with defined emotion classes. The hashtags in the tweet represent part of the content of those tweets. As a result, only tweets with hashtags at the end have been gathered [14].

One of the advantages of the hashtag data extraction method is the collection of a large number of tweets with different hashtags of emotions without manual interference. Another advantage of this approach is that instead of relying on those responsible for tweeting interpretation, we can have direct access to the emotions that the author intended to express, as the author's feeling is likely to be misinterpreted by the interpreter a lot [14].

*B. Results and Discussions*

To analyze the results, first various trainings have been done on the proposed network with different parameters and the best weight is applied as the network weight to measure the test data.

Information about three training experiments is given in the Table II.

TABLE II
PARAMETERS IN DIFFERENT TRAINING EXPERIMENTS

|  | Maximum Repetitive Words | Maximum Sentence Length | Evaluation Separator | Dimension Embedding Layer |
|---|---|---|---|---|
| First Experiment | 40000 | 30 | 0.2 | 200 |
| Second Experiment | 40000 | 15 | 0.2 | 200 |
| Third Experiment | 10000 | 30 | 0.2 | 200 |

After training the network using the parameters of Table II, evaluation data confusion matrix has been shown to evaluate the proposed model and to check it with test data confusion matrix in order to detect network non-overfitting during training as shown in Fig. 8.

To determine the presence or absence of an overfitting problem, it is necessary to compare the confusion matrix of the evaluation data in Fig. 8 with the confusion matrix of the test data in Fig. 9.

|  | Happy-Active | Happy-Inactive | Unhappy-Active | Unhappy-Inactive |
|---|---|---|---|---|
| Happy-Active | 0.96 | 0.01 | 0.02 | 0.01 |
| Happy-Inactive | 0.03 | 0.93 | 0.02 | 0.02 |
| Unhappy-Active | 0.01 | 0.01 | 0.96 | 0.02 |
| Unhappy-Inactive | 0.02 | 0.01 | 0.03 | 0.94 |

Fig. 8 (a) Confusion Matrix of the Evaluation Data in The First Experiment

|  | Happy-Active | Happy-Inactive | Unhappy-Active | Unhappy-Inactive |
|---|---|---|---|---|
| Happy-Active | 0.93 | 0.02 | 0.03 | 0.03 |
| Happy-Inactive | 0.04 | 0.89 | 0.04 | 0.03 |
| Unhappy-Active | 0.02 | 0.01 | 0.93 | 0.04 |
| Unhappy-Inactive | 0.03 | 0.01 | 0.06 | 0.90 |

Fig. 8 (b) Confusion Matrix of the Evaluation Data in The Second Experiment

|  | Happy-Active | Happy-Inactive | Unhappy-Active | Unhappy-Inactive |
|---|---|---|---|---|
| Happy-Active | 0.95 | 0.01 | 0.02 | 0.02 |
| Happy-Inactive | 0.04 | 0.91 | 0.02 | 0.03 |
| Unhappy-Active | 0.01 | 0.01 | 0.95 | 0.03 |
| Unhappy-Inactive | 0.02 | 0.01 | 0.05 | 0.92 |

Fig. 8 (c) Confusion Matrix of the Evaluation Data in The Third Experiment

TABLE III
EVALUATION CRITERIA OBTAINED IN THREE EXPERIMENTS

| Emotion Classes | First Experiment | | | Second Experiment | | | Third Experiment | | |
|---|---|---|---|---|---|---|---|---|---|
|  | Prec. | Rec. | FM | Prec. | Rec. | FM | Prec. | Rec. | FM |
| Happy-Active | 94 | 96 | 95 | 92 | 93 | 92 | 94 | 95 | 94 |
| Happy-Inactive | 96 | 93 | 94 | 95 | 89 | 91 | 96 | 91 | 93 |
| Unhappy-Active | 95 | 96 | 96 | 90 | 93 | 92 | 92 | 95 | 93 |
| Unhappy-Inactive | 94 | 94 | 96 | 90 | 90 | 90 | 92 | 92 | 92 |

|  | Happy-Active | Happy-Inactive | Unhappy-Active | Unhappy-Inactive |
|---|---|---|---|---|
| Happy-Active | 0.94 | 0.02 | 0.02 | 0.02 |
| Happy-Inactive | 0.04 | 0.91 | 0.02 | 0.03 |
| Unhappy-Active | 0.02 | 0.01 | 0.92 | 0.05 |
| Unhappy-Inactive | 0.03 | 0.02 | 0.05 | 0.91 |

Fig. 9 (a) Confusion Matrix of the Test Data in The First Experiment

|  | Happy-Active | Happy-Inactive | Unhappy-Active | Unhappy-Inactive |
|---|---|---|---|---|
| Happy-Active | 0.93 | 0.02 | 0.02 | 0.03 |
| Happy-Inactive | 0.04 | 0.90 | 0.03 | 0.03 |
| Unhappy-Active | 0.02 | 0.01 | 0.92 | 0.04 |
| Unhappy-Inactive | 0.03 | 0.02 | 0.06 | 0.90 |

Fig. 9 (b) Confusion Matrix of the Test Data in The Second Experiment

|  | Happy-Active | Happy-Inactive | Unhappy-Active | Unhappy-Inactive |
|---|---|---|---|---|
| Happy-Active | 0.93 | 0.02 | 0.02 | 0.02 |
| Happy-Inactive | 0.04 | 0.90 | 0.03 | 0.02 |
| Unhappy-Active | 0.02 | 0.01 | 0.93 | 0.05 |
| Unhappy-Inactive | 0.03 | 0.01 | 0.05 | 0.91 |

Fig. 9 (c) Confusion Matrix of the Test Data in The Third Experiment

TABLE IV
EVALUATION CRITERIA OF THE TEST DATA OF THREE EXPERIMENTS

| Emotion Classes | First Experiment | | | Second Experiment | | | Third Experiment | | |
|---|---|---|---|---|---|---|---|---|---|
|  | Prec. | Rec. | FM | Prec. | Rec. | FM | Prec. | Rec. | FM |
| Happy-Active | 92 | 94 | 93 | 91 | 93 | 92 | 92 | 93 | 93 |
| Happy-Inactive | 93 | 91 | 92 | 94 | 90 | 92 | 94 | 90 | 92 |
| Unhappy-Active | 92 | 92 | 92 | 91 | 92 | 91 | 91 | 93 | 92 |
| Unhappy-Inactive | 90 | 91 | 90 | 90 | 90 | 90 | 91 | 91 | 91 |

The resulting confusion matrix is shown in Fig. 9. A quick glance at the confusion matrix reveals that by comparing the three experiments, it can be seen that the F-score criterion in the second experiment is lower than the first and third experiments, and this shows that when we reduce the maximum parameter value of the sentence length- that is the lower number of tokens is considered per sentence and if this value is less to some extent, i.e., more tokens are discarded- the obtained results will get a lower F-score. Also, the comparison of the first and third experiments does not show any change, especially in the evaluation criteria.

*C. Evaluation of All Tweets*

Fig. 10 shows a general evaluation of all the tweets extracted from the Twitter social network, along with a prediction of the emotion classes of those tweets, extracted from the code in the Python language using Keras [30] and sklearn packages [31]. Due to the large number of tweets, a small part of it is included in the article randomly. These tweets are pre-processed tweets, and there is a predicted class below each tweet.

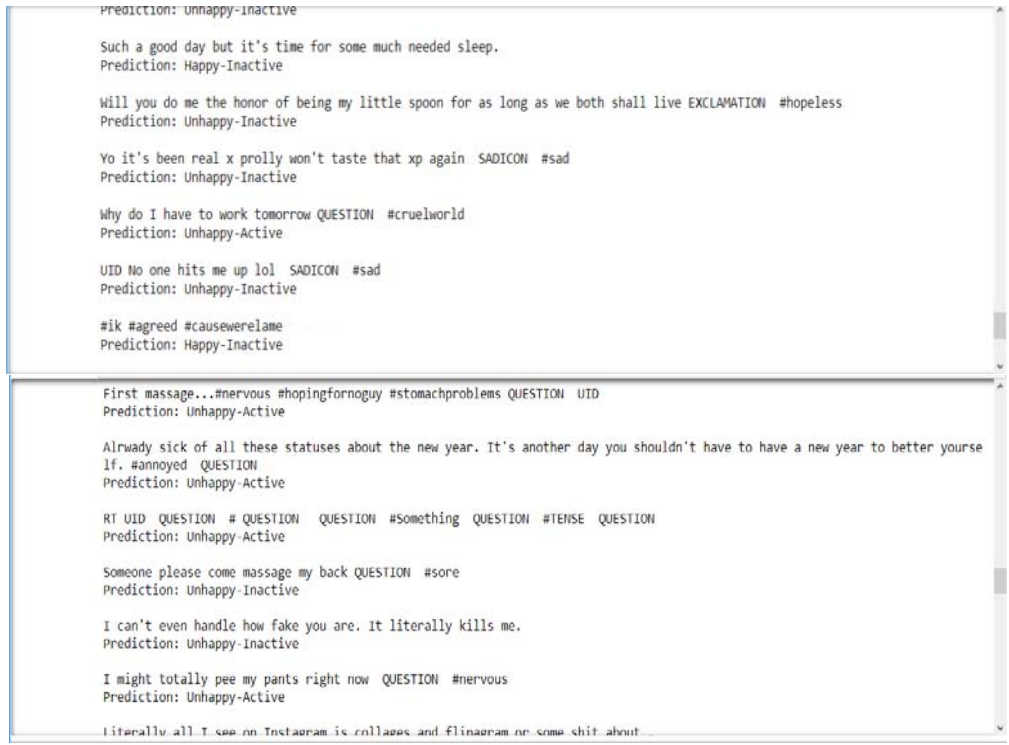

Fig. 10 Part of the class prediction on all tweets

## VI. Conclusions

This article has presented the proposed method of a combined model from two deep neural networks of long short term memory and convolution in order to classify emotional states in the social network Twitter. And an attempt was made to teach the deep neural network used in this study and to evaluate it with test data in order to improve the classification of emotions in the data set. The most important goal that was envisaged in this work was to achieve an acceptable accuracy in classifying tweets in terms of emotional states and increase of the previous work accuracy. In the proposed method, we achieved an average accuracy of 93%, which we can find an acceptable amount of this accuracy by comparing previous work.

In the course of this research, some cases were observed that can be considered in future work, including the implementation of emotion classification in other languages, such as Persian, despite its great linguistic complexity, the recognition and classification of emotions in respective areas, such as following the flow of ideas propagation on social media and illustrating the propagation flow, as well as classification of ironic and slang content.